\pdfoutput=1
\documentclass[11pt]{article}
\usepackage[preprint]{acl}
\usepackage{longtable}
\usepackage{geometry}
\usepackage{graphicx}
\usepackage{adjustbox}
\usepackage{amssymb}
\usepackage{subfigure}
\usepackage{subcaption}
\usepackage{array,booktabs,ragged2e}
\usepackage{multirow}
\usepackage{multicol}
\usepackage{times}
\usepackage{latexsym}
\usepackage{arydshln}
\usepackage{hyperref}
\usepackage{pifont}% http://ctan.org/pkg/pifont
\usepackage{float}
\usepackage{bm}
\usepackage{microtype}
\usepackage{inconsolata}
\usepackage{xcolor}
\usepackage{xspace}

\usepackage{placeins}
\usepackage{afterpage}
\usepackage{amsfonts}
\usepackage{amsmath}

\usepackage[T1]{fontenc}
\usepackage[utf8]{inputenc}

\definecolor{black}{HTML}{000000}
\definecolor{Ind}{HTML}{D62F2F}
\definecolor{US}{HTML}{485694}
\definecolor{Trans}{HTML}{485694}
\definecolor{Multi}{HTML}{D48B0F}

\newcommand{\IndEng}{en-IN\xspace}
\newcommand{\USEng}{en-US\xspace}
\newcommand{\Trans}{IN-TR\xspace}
\newcommand{\Multi}{IN-MV\xspace}

\newcommand{\diff}[1]{\textcolor{Ind}{\small{(#1)}}}
\newcommand{\improve}[1]{\textcolor{Trans}{\small{(#1)}}}

\newcommand{\Tar}{{\text{[MASK]}}_{\text{US}}}
\newcommand{\Dial}{{\text{[MASK]}}_{\text{X}}}

\newcommand{\mistral}{\textsc{mistral}\xspace}
\newcommand{\gemma}{\textsc{gemma}\xspace}
\newcommand{\ours}{\textsc{LoRDD}\xspace}

\newcommand{\googlogo}{\raisebox{5pt}{\includegraphics[scale=0.09]{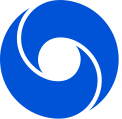}}}
\newcommand{\unswlogo}{\raisebox{3.4pt}{\includegraphics[scale=0.120]{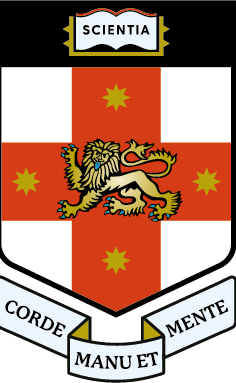}}}

\title{Predicting the Target Word of Game-playing Conversations using a Low-Rank Dialect Adapter for Decoder Models}
\author{
Dipankar Srirag\unswlogo\quad Aditya Joshi\unswlogo\quad Jacob Eisenstein\googlogo\\
\unswlogo University of New South Wales, Sydney\quad\googlogo Google DeepMind\\
  \texttt{\{d.srirag, aditya.joshi\}@unsw.edu.au}\quad\texttt{jeisenstein@google.com}
  }

\begin{document}
\maketitle
\begin{abstract}
Dialect adapters that improve the performance of LLMs for NLU tasks on certain sociolects/dialects/national varieties (`dialects' for the sake of brevity) have been reported for encoder models. In this paper, we extend the idea of dialect adapters to decoder models in our architecture called \ours. Using MD-3, a publicly available dataset of word game-playing conversations between dialectal speakers, our task is Target Word Prediction (TWP) from a masked conversation. \ours combines task adapters and dialect adapters where the latter employ contrastive learning on pseudo-parallel conversations from MD-3. Our experiments on Indian English and Nigerian English conversations with two models (\mistral and \gemma) demonstrate that \ours outperforms four baselines on TWP. Additionally, it significantly reduces the performance gap with American English, narrowing it to 12\% and 5.8\% for word similarity, and 25\% and 4.5\% for accuracy, respectively. The focused contribution of \ours is in its promise for dialect adaptation of decoder models using TWP, a simplified version of the commonly used next-word prediction task.
\end{abstract}
\section{Introduction}\label{sec:intro}

\begin{figure}[t!]
    \centering
    \includegraphics[width=1\linewidth]{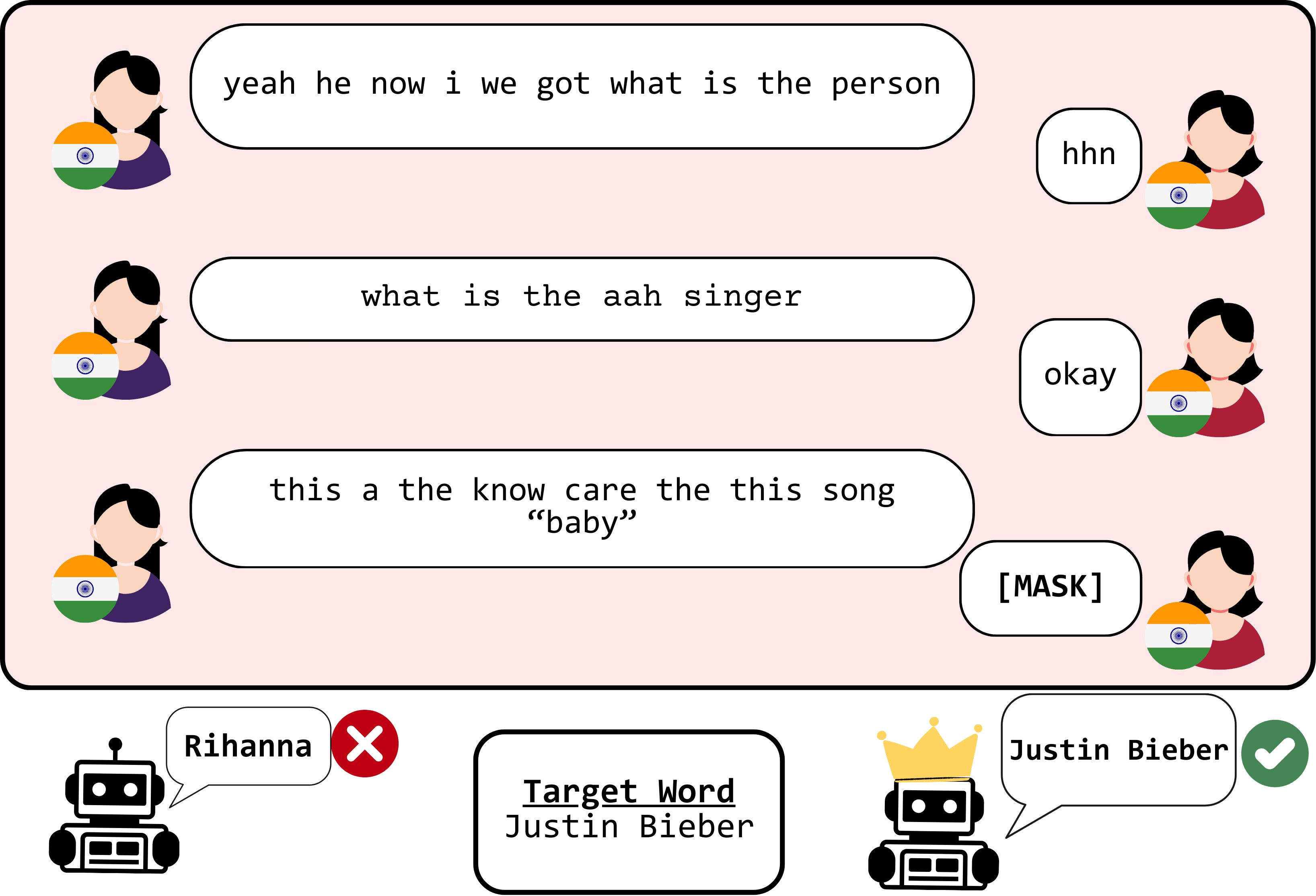}
    \caption{Illustrative example of Target Word Prediction on an en-IN conversation. The inaccurate output from the in-dialect fine-tuned model (left) is corrected by the model trained using \ours (right).}
    \label{fig:illustration}
\end{figure}

Dialect adaptation of language models refers to approaches that improve their performance for different dialects of a language~\cite{joshi2024natural}. Past work proposes dialect adaptation for encoder models~\cite{held-etal-2023-tada, xiao-etal-2023-task} or encoder-decoder models~\cite{liu-etal-2023-dada}. This paper extends it to decoder models, via a novel architecture called \textbf{Lo}w-\textbf{R}ank \textbf{D}ialect robustness for \textbf{D}ecoder Models (\ours). To demonstrate the effectiveness of \ours, we use MD-3~\cite{eisenstein2023md3}, a dataset of manually transcribed dialectal dialogues between speakers of either Indian English (en-IN) or Nigerian English (en-NG) or US English (en-US) playing the word-guessing game of taboo\footnote{In a game of taboo, a describer must get a guesser to guess a target word without using a set of words known as taboo words.}. We select MD-3 conversations where the guesser correctly identifies the target word/phrase (`target word' for the sake of brevity) and mask the target word (using \text{[MASK]}; as shown in Figure~\ref{fig:illustration}). Our task then is to predict the target word in a masked conversation, \textit{i.e.}, target word prediction (TWP). TWP represents a simplified version of next-word generation utilised by decoder models. Since decoder models are adept in tasks involving causal language modeling, TWP is a reasonable task choice. Upon observing that the TWP performances for en-IN and en-NG are lower than those of en-US, the objective of \ours is to improve the TWP performances for en-IN and en-NG.
\ours employs a combination of two LoRA-based~\cite{hu2021loralowrankadaptationlarge} adapters. The first is a task-specific adapter that uses instruction fine-tuning~\cite{wei2022finetunedlanguagemodelszeroshot} on an augmented set of en-US and en-IN/en-NG conversations. The second is a dialect adapter that uses contrastive learning on a pseudo-parallel corpus between en-US and en-IN/en-NG conversations about a specific target word. We release the code for training \ours adapters on \hyperlink{https://github.com/dipankarsrirag/lordd}{Github}.

Our work is novel in two ways: (A) \ours is the first methodology for dialect adaptation of decoder models, and outperforms one in-dialect and three cross-dialect baselines, (B) We leverage an existing dataset MD-3 to create a pseudo-parallel corpus of natural dialectal conversations, as opposed to past work that relies on synthetically transformed dialectal corpora.

% Dialects are more visible in conversations between speakers of that dialect than in the written form of the language. (Which is what the previous study used)
% \paragraph{Contributions}
% We propose \ours as a novel framework to improve dialect robustness for decoder models and show its empirical effectiveness in bridging the performance gap on en-IN conversations for the Target Word Prediction task. 
% \paragraph{Introduce the problem} 
% Large Language Models (LLMs;~\citealp{gemmateam2024gemma, jiang2023mistral, openai2024gpt4}) were pre-trained on large textual corpora. Although the pre-training corpora are diverse in regards to the domain, they often suffer from being concentrated in languages and more specifically majority dialects~\cite{???}. This leads to a disparity in the performances of these LLMs on several NLP tasks that involve texts from minority dialects~\cite{joshi2024natural}.
% \paragraph{Overview of previous works} Three sentences summarizing dialect adaptation work. Must include TADA.

%Assuming dialects are an alternate form of a language, we follow the work done in the multilingual alignment of closely related languages by adapting pre-trained models to a dialect~\cite {???}(multi-dialectal alignment). 

% \paragraph{Task} What is the task? How do we pick the task? Why not other tasks?

%INTRODUCE TARGET WORD PREDICTION in THREE SENTENCES.
\section{Architecture of \ours}\label{sec:method}
\begin{figure*}[h!]
    \centering
    \includegraphics[width=0.75\linewidth]{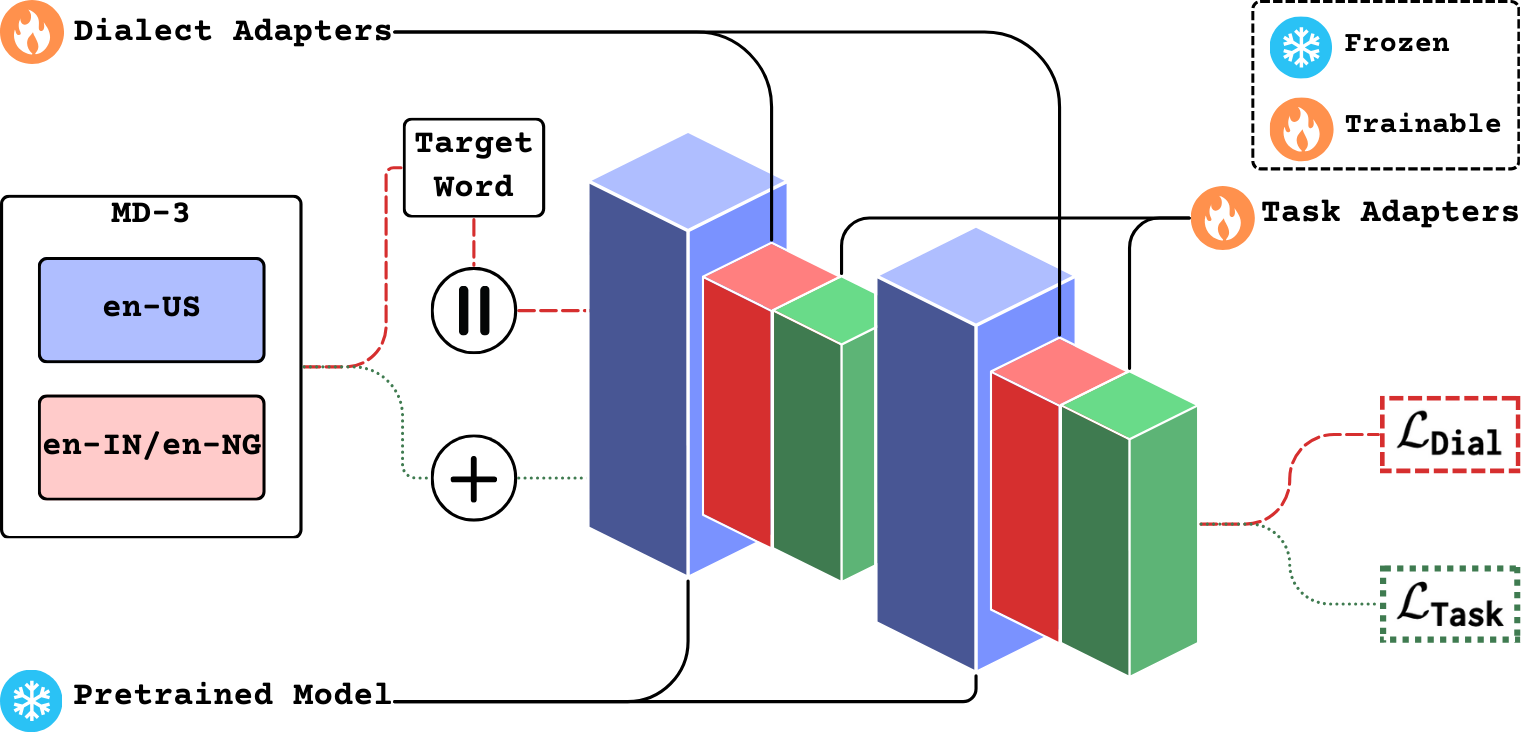}
    \caption{Architecture of \ours.}
    \label{fig:lordd}
\end{figure*}

The architecture of \ours employs two parameter-efficient adapters: task adapter and dialect adapter, as shown in Figure~\ref{fig:lordd}. 
\subsection{Task Adapter}
We define $\mathbf{x}$ and $\mathbf{t}$ as lists of tokens in the masked conversation and the target word respectively. For a batched input of $N$ pairs of masked conversations and corresponding target words, we train the task adapters to output the correct target word using maximum likelihood estimation -- a standard learning objective for causal language modeling~\cite{jain-etal-2023-contraclm}.
 \begin{align*}
 % \scriptstyle
    \mathcal{L}_{\texttt{\textbf{Task}}} = - \frac{1}{N}\sum_{j=1}^{N}\left\{\sum_{i=|\mathbf{x}^j|+1}^{|\mathbf{x}^j|+|\mathbf{t}^j|} \log p(\mathbf{x}_i^j | \mathbf{x}_{< i}^j) \right\}\nonumber
\end{align*}
Here, $\mathbf{x}_{< i}^j = [\mathbf{x}_1^j, \ldots, \mathbf{x}_{i-1}^j]$ denotes the subsequence before $\mathbf{x}_{i}^j$ and $|\cdot|$ is the number of tokens. 

\subsection{Dialect Adapter}

To train the dialect adapter, we use a pseudo-parallel corpus between en-US and en-IN/en-NG conversations. This corpus consists of both positive and negative pairs of masked conversations. We consider a masked conversation pair as a positive example if both conversations pertain to the same target word, and a negative example if they pertain to a different target word. We then perform contrastive learning between the frozen representation of the masked en-US conversation ($\Tar$) and the trainable representation of the masked en-IN/en-NG conversation ($\Dial$), using cosine embedding loss. This allows the adapters to learn from both positive and negative examples present in the pseudo-parallel corpus.

\begin{align*}
    \small
    \mathcal{L}_{\texttt{\textbf{Dial}}} = 
    \begin{cases}
    \text{1 - sim}(\Tar, \Dial)\text{; y = 1} \\
    \max{\left(\text{0, sim}(\Tar, \Dial)\text{ - d}\right)}\text{; y = -1} \\ 
    \end{cases}\nonumber
\end{align*}

Here, X represents dialect in focus (either en-IN or en-NG), sim($\cdot$) calculates the cosine similarity, `d' is the margin, and `y' is the label (1 for a positive example, and -1 otherwise).

In contrast to the task adapter, the dialect adapter is trained to output standard dialect representations for an input text. Hence, \ours stacks the task adapter on top of the dialect adapter (as shown in Figure~\ref{fig:lordd}), allowing the models to predict the target word as required for TWP.

% \begin{align}
%     \cos \varphi = \frac{\langle \Tar \cdot \Dial \rangle}{||\Tar||\times||\Dial||}
%     \nonumber
% \end{align}
\section{Experiment Setup}\label{sec:setup}
% Negative samples are randomly selected
We experiment with two open-weight decoder models namely, Mistral-7B-Instruct-v0.2 (\mistral;~\citealp{jiang2023mistral7b}) and Gemma2-9B-Instruct (\gemma;~\citealp{gemmateam2024gemma2improvingopen}). \ours is trained as follows:
% We train \ours in two phases:
\begin{itemize}
    \item The task adapter is trained by fine-tuning the model for 20 epochs, with a batch size of 32, Paged 8-bit AdamW~\cite{dettmers20228bit} as the optimiser and learning rate of 2e-4.
    \item To train the dialect adapter, we perform contrastive learning for 10 epochs, with a batch size of 8, AdamW as the optimiser, a learning rate of 2e-5, and a margin of 0.25.
\end{itemize}

We inject adapter matrices at all linear layers, as recommended by~\citet{dettmers2023qlora}. Training either adapter for a single experiment takes approx. 25 minutes on an A100 GPU.
\begin{table}[h!]
     \begin{adjustbox}{width=0.85\linewidth,center}
     \scriptsize{
         \begin{tabular}{m{4em}ccc}
         \toprule
         \centering Subset & Train & Valid & Test\\\midrule[\heavyrulewidth]
         \centering en-US & 62 & 41 & 311\\
         \centering en-IN & 31 & 21 & 160\\
         \centering en-NG & 38 & 25 & 194\\\hdashline
         \centering IN-MV & 57 & 39 & 296\\
         \centering NG-MV & 57 & 39 & 296\\\hdashline
         \centering IN-TR & 25 & 17 & 132\\
         \bottomrule
         \end{tabular}
         }
     \end{adjustbox}
     \caption{\label{tab:stats}
     Data statistics.}
 \end{table}
We compare \ours with one in-dialect and three cross-dialect baselines. The in-dialect baseline involves fine-tuning a model on the training set of en-IN/en-NG. The cross-dialect baselines are:%\\
\paragraph{en-US} Fine-tune the model on train set of en-US.%\\
\paragraph{IN-MV/NG-MV} We use Multi-VALUE~\cite{ziems2023multi} to transform en-US conversations into en-IN. IN-MV is fine-tuned on these synthetically created conversations.%\\
\paragraph{IN-TR} We  prompt GPT-4 Turbo~\cite{openai2024gpt4technicalreport} to \emph{transform} en-IN by removing dialectal information, resulting in IN-TR, and use it to fine-tune a model.\\
\textbf{Note:} We do not perform similar transformations on the en-NG subset due to the high API pricing at the time of writing. 

We consider the in-dialect fine-tuned model as a strong baseline, while cross-dialect models are weak baselines. We compare all baselines and \ours with in-dialect fine-tuned models on en-US conversations, which serves as our skyline result.
 \begin{table}[h!]
     \begin{adjustbox}{width=\linewidth,center}
     \scriptsize{
         \begin{tabular}{m{6.68em}ccc}
         \toprule
         \centering $\textbf{||}_{\text{Corpus}}$ & Samples & Positive & Negative\\\midrule[\heavyrulewidth]
         \centering en-US || en-IN & 144  & 11 & 133\\
         \centering en-US || en-NG & 168  & 13 & 155\\\hdashline
         \centering en-US || IN-MV & 197  & 97 & 100\\
         \centering en-US || NG-MV & 197  & 97 & 100\\\hdashline
         \centering en-IN || IN-TR & 142 & 42 & 100\\
         \bottomrule
         \end{tabular}
         }
     \end{adjustbox}
     \caption{\label{tab:par_stats}
     Data statistics of the pseudo-parallel corpus.}
 \end{table}

Tables~\ref{tab:stats} and ~\ref{tab:par_stats} report the statistics of the extended MD-3 dataset and the pseudo-parallel corpus respectively. Additional details including prompt used to create TR-IN and corpus examples are in Appendix~\ref{sec:data_con}. All evaluations are on the test set of the en-IN or en-NG subsets for the baselines and \ours, and on the test set of the en-US dataset for the skyline. We report two metrics: (a) Similarity (average cosine similarity between the Sentence-BERT~\cite{reimers2019sentencebert} embeddings of the reference and generated target word); and (b) Accuracy (the proportion of conversations where the model generates the correct target word).

\section{Evaluation}\label{sec:results}
Our results address three questions: (a) What is the current gap in the task performance between en-US and en-IN/en-NG?; (b) How well does \ours help bridge the gap?; (c) How essential is each component in \ours to bridge the gap?

\begin{table*}[ht!]
    \centering
    \begin{adjustbox}{width=1.0\linewidth,center}
        \setlength{\cmidrulekern}{0.25em}
        \begin{tabular}{m{8.7em}m{6.2em}cccccc}
        \toprule
         \multirow{2}{8.7em}{\centering Method} & \multirow{2}{6.2em}{\centering Training Data} & \multicolumn{2}{c}{\mistral} & \multicolumn{2}{c}{\gemma}&\multicolumn{2}{c}{$\mu$}\\\cmidrule(lr){3-4}\cmidrule(lr){5-6}\cmidrule(lr){7-8}
         & & Similarity & Accuracy & Similarity & Accuracy & Similarity & Accuracy\\\midrule[\heavyrulewidth]
         \centering Skyline %
             & \centering en-US & 64.7 & 44.3 & 69.7 & 45.3 &\diff{0.0} 67.2 \improve{27.3} &\diff{0.0} 44.8 \improve{64.7}\\\midrule[\heavyrulewidth]
        \multicolumn{8}{c}{(a) Tested on en-IN}\\\midrule[\heavyrulewidth]
         \centering In-dialect baseline  %
             & \centering en-IN & 51.0 & 24.4 & 54.6 & 30.0 & \diff{27.3} 52.8 \improve{0.0} & \diff{64.7} 27.2 \improve{0.0}\\\hdashline
         \multirow{3}{8.7em}{\centering Cross-dialect baseline} %
             & \centering en-US & 54.6 & 25.6 & 61.3 & 35.0 & 58.0 & 30.3\\
             & \centering IN-MV & 52.4 & 24.4 & 58.2 & 30.0 & 55.3 & 27.2\\
             & \centering IN-TR & 50.4 & 24.3 & 53.0 & 26.9 & 52.7 & 25.6\\\hdashline
         \centering \ours %
             & \centering en-US + en-IN & \textbf{55.9} & \textbf{30.0} & \textbf{63.9} & \textbf{41.3} & \diff{12.0} \textbf{59.9} \improve{13.4} & \diff{25.0} \textbf{35.7} \improve{28.1}\\\midrule[\heavyrulewidth]
        \multicolumn{8}{c}{(b) Tested on en-NG}\\\midrule[\heavyrulewidth]
        \centering In-dialect baseline  %
            & \centering en-NG & 53.0 & 27.2 & 60.9 & 35.3 & \diff{17.9} 57.0 \improve{0.0} & \diff{43.1} 31.3 \improve{0.0}\\\hdashline
        \multirow{2}{8.7em}{\centering Cross-dialect baseline} %
            & \centering en-US & 58.9 & 31.4 & 62.8 & 40.7 & 60.9 & 36.1\\
            & \centering NG-MV & 55.7 & 28.4 & 61.4 & 38.6 & 58.9 & 33.5\\\hdashline
        \centering \ours %  
            &\centering en-US + en-NG & \textbf{62.4} & \textbf{40.5} & \textbf{64.5} & \textbf{43.2} & \diff{5.8} \textbf{63.5} \improve{11.4} & \diff{4.5} \textbf{41.9} \improve{33.8}\\
         % \multicolumn{2}{c}{\texttt{Ours}}& 55.9 & 30.0 & 62.8 & 38.8 & 7.9 &  10.4\\\hdashline
        \bottomrule
        %& 13.7 & 19.9
        \end{tabular}
    \end{adjustbox}
    \caption{\label{tab:baselines}Performance comparison between the skyline, baselines and \ours on TWP. For each model, we report  Similarity and Accuracy when tested on (a) en-IN and (b) en-NG. $\mu$ is the average of the metrics across both evaluation models. \ours (represented in \textbf{bold}) improves the performance on all baselines. The percentage improvement over the in-dialect baseline and the percentage degradation compared to the skyline are shown in \improve{number} and \diff{number} respectively.}
\end{table*}
Table~\ref{tab:baselines} compares the performance of \ours with the baselines and the skyline. On the similarity and accuracy, \ours achieves average scores of 59.9 and 35.7, respectively, when evaluated on en-IN, and 63.5 and 41.9, respectively, when evaluated on en-NG. On average, \ours improves on the performances of the en-IN in-dialect baseline by 13.4\% on similarity and 28.1\% on accuracy. Similarly, it improves on the en-NG in-dialect baseline by 11.4\% on similarity and 33.8\% on accuracy. 

As expected, the skyline achieves the highest performance for the task. However, \ours significantly narrows the initial performance gaps. For en-IN, the gap in similarity is reduced from 27.3\% to 12\%, and the gap in accuracy is reduced from 64.7\% to 25\%. For en-NG, the gap in similarity is reduced from 17.9\% to 5.8\%, and the gap in accuracy is reduced from 43.1\% to 4.5\%.
\begin{table*}[h!]
    \centering
    \begin{adjustbox}{width=1.0\linewidth,center}
        \begin{tabular}{m{4.2em}m{6.68em}m{6.68em}cccccc}
        \toprule
        \multirow{2}{4.2em}{\centering Method} & \multirow{2}{6.68em}{\centering Training Data} & \multirow{2}{6.68em}{\centering $\textbf{||}_{\text{Corpus}}$}& \multicolumn{2}{c}{\mistral} & \multicolumn{2}{c}{\gemma} & \multicolumn{2}{c}{$\mu$}\\\cmidrule(lr){4-5}\cmidrule(lr){6-7}\cmidrule(lr){8-9}
          & & & Similarity & Accuracy & Similarity & Accuracy & Similarity & Accuracy\\\midrule[\heavyrulewidth]
        \multicolumn{9}{c}{(a) Tested on en-IN}\\\midrule[\heavyrulewidth]
         \centering \ours %
         & \centering en-US + en-IN & \centering en-US || en-IN & \textbf{55.9} & \textbf{30.0} & \textbf{63.9} & \textbf{41.3} & \textbf{59.9} & \textbf{35.7}\\\hdashline
         \multirow{2}{4.2em}{\centering $\leftrightarrow \textbf{||}_{\text{Corpus}}$} %
         & \centering en-US + en-IN & \centering en-US || IN-MV & 55.6 & 28.1 & 62.0 & 37.5 & 58.8 \diff{1.1} & 32.8 \diff{2.9}\\
          & \centering en-US + en-IN & \centering en-IN || IN-TR & 54.9 & 27.5 & 62.8 & 38.8 & 58.9 \diff{1.0} & 33.2 \diff{2.5}\\\hdashline
         \multirow{3}{4.2em}{\centering $-\mathcal{L}_{\texttt{\textbf{Dial}}}$} %
         & \centering en-US + en-IN & \multirow{3}{6.68em}{\centering Not Used} & 54.4 & 26.9 & 62.3 & 37.5 & 58.4 \diff{1.5} & 32.2 \diff{3.5}\\
         & \centering en-IN + IN-MV &  & 51.6 & 23.1 & 57.1 & 31.9 & 54.4 \diff{5.5} & 27.5 \diff{8.2}\\
         & \centering en-IN + IN-TR &  & 44.8 & 18.1 & 57.5 & 28.8 & 51.2 \diff{8.7} & 23.5 \diff{12.2}\\\midrule[\heavyrulewidth]
         \multicolumn{9}{c}{(b) Tested on en-NG}\\\midrule[\heavyrulewidth]
         \centering \ours %
            & \centering en-US + en-NG & \centering en-US || en-NG & \textbf{62.4} & \textbf{40.5} & \textbf{64.5} & \textbf{43.2} & \textbf{63.5} & \textbf{41.9}\\\hdashline
         \centering $\leftrightarrow \textbf{||}_{\text{Corpus}}$ %
            & \centering en-US + en-NG & \centering en-US || NG-MV & 60.4 & 35.6 & 61.9 & 38.5 & 61.2 \diff{2.3} & 37.1 \diff{4.8}\\\hdashline
         \multirow{2}{4.2em}{\centering $-\mathcal{L}_{\texttt{\textbf{Dial}}}$} %
             & \centering en-US + en-NG & \multirow{2}{6.68em}{\centering Not Used} & 61.3 & 39.7 & 62.4 & 38.1 & 61.9 \diff{1.6} & 38.9 \diff{3.0}\\
             & \centering en-IN + NG-MV &  & 58.6 & 33.6 & 60.7 & 33.1 & 59.7 \diff{3.8} & 33.4 \diff{8.5}\\
        \bottomrule
        \end{tabular}
    \end{adjustbox}
    \caption{\label{tab:ablation}
    Ablation on \ours based on parallel corpus ($\leftrightarrow \textbf{||}_{\text{Corpus}}$), dialect adapter ($\mathcal{L}_{\texttt{\textbf{Dial}}}$) and data augmentation. For each model, we report Similarity and Accuracy when tested on (a) en-IN and (b) en-NG. The best performance is shown in \textbf{bold}. $\mu$ is the average of the metrics across both models. The degradation on the ablations compared to \ours is shown in \diff{number}.
    }
\end{table*}

Table~\ref{tab:ablation} shows the results from an ablation study that evaluates both adapters in \ours. We compare \ours with three variants: (a) the dialect adapter trained on other parallel corpora, (b) \ours without the dialect adapter, within which we also compare, (c) the task adapters trained on other augmented data. Compared to \ours, all other variants report a degradation in their performances. Training the dialect adapter on synthetic parallel corpora (en-US || IN-MV, en-IN || IN-TR and en-US || NG-MV) results in degradation ranging from 1.0 to 2.3 on similarity and 2.5 to 4.8 on accuracy. Removing the dialect adapter results in a further degradation ranging from 1.5 to 8.7 on similarity and 3.0 to 12.2 on accuracy. The worst-performing variants are the models that only train the task adapter on synthetically augmented data (en-US + IN-MV, en-IN + IN-TR and en-IN + NG-MV). While the degraded performances of these models show the importance of the dialect adapter, the lower performances on variants involving synthetic conversations further solidify the use of natural conversations in \ours. We provide additional results, such as ablations on proportion of conversations in augmented data, in Appendix~\ref{sec:add_ab}.

\begin{figure}[t!]
    \centering
    \includegraphics[width=\linewidth]{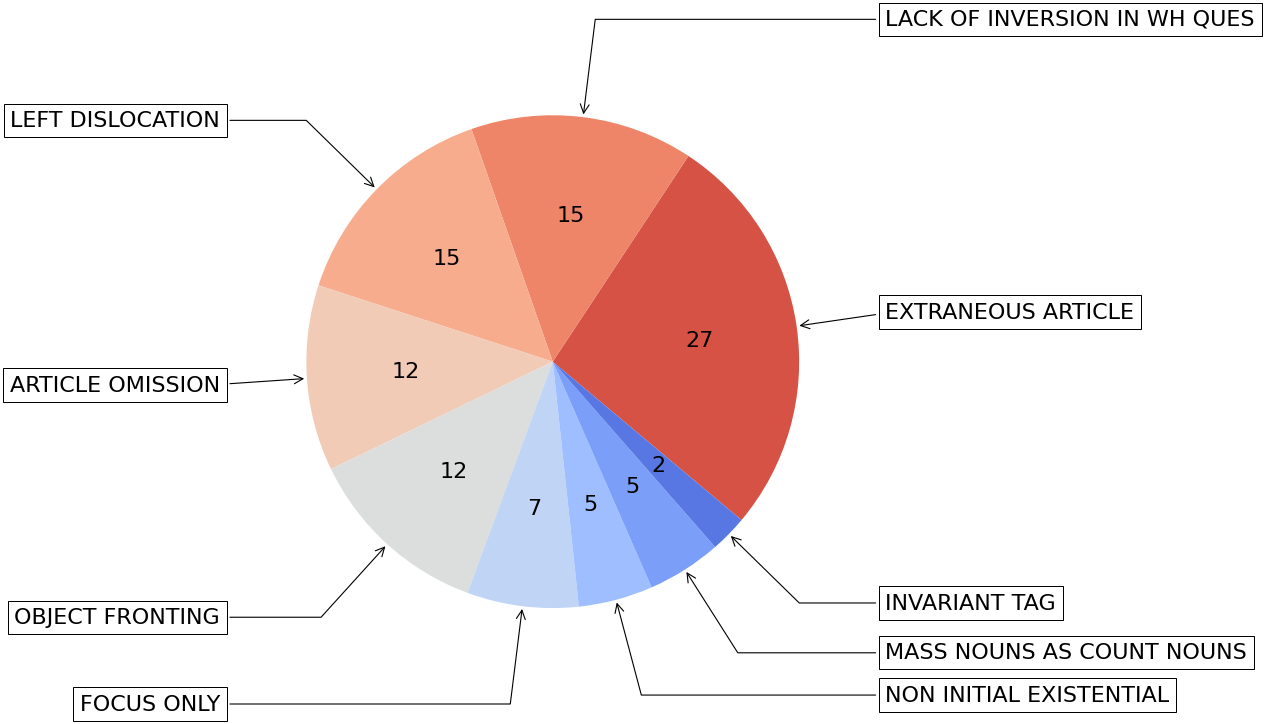}
    \caption{Percentage count of dialect features in erroneous instances from \ours.}
    \label{fig:err_pie}
\end{figure}
Finally, we manually analyse erroneous en-IN instances from \ours, and categorise them into types of en-IN dialect features given by~\citet{lange2012syntax} and~\citet{demszky-etal-2021-learning}. Figure~\ref{fig:err_pie} shows that \textsc{extraneous article} (\emph{``It's \underline{a} one word''}) is the most common feature associated with these conversations. The definitions of all identified dialect features with examples are in Table~\ref{tab:dial_feat}.\\
\textbf{Note:} We do not perform error analysis for en-NG instances due to lack of similar labelled features for the dialect. 

% Finally, we manually analyse erroneous instances from \ours, and categorise them into types of dialect features as given by~\citet{lange2012syntax} and~\citet{demszky-etal-2021-learning}. Figure~\ref{fig:err_pie} shows that \textsc{extraneous article} (\emph{``It's \underline{a} one word''}) and \textsc{lack of inversion in wh-questions} (\emph{``what \underline{we can} see in the rivers?''}) are the most common features associated with these conversations. The definitions of all dialectal features with examples are in Appendix~\ref{sec:dial_feat}.

\section{Related Work}\label{sec:related}
Language technologies need to be equitable to dialects/sociolects/national varieties~\cite{joshi2024natural, blodgett-etal-2020-language}. Dialect adaptation involves strategies to improve the performance of non-mainstream dialects. These strategies range from introducing dialectal information at the pre-training phase~\cite{sun-etal-2023-dialect} to adapter-based approaches. Adapters are explored to be viable and efficient in improving dialect robustness~\cite{liu-etal-2023-dada} or cross-lingual transfer~\cite{pfeiffer2020mad}. In particular, we derive from this line of work by training a low-rank dialect adapter like~\citet{xiao-etal-2023-task} using a contrastive learning objective like~\citet{held-etal-2023-tada}. While past approaches adapt encoder models, we distinguish ourselves by proposing \ours as an architecture to adapt decoder models. Similarly, past work uses frameworks like VALUE~\cite{ziems-etal-2022-value} and Multi-VALUE~\cite{ziems2023multi} to create synthetic dialectal variants of standard US English benchmarks. In contrast, we use a pseudo-parallel corpus of naturally occurring dialectal conversations from MD-3~\cite{eisenstein2023md3}. Our task of target word prediction is closely similar to~\citet{chalamalasetti-etal-2023-clembench}, who generate word game conversations using LLMs and evaluate their ability to predict the target word. Target word prediction is also utilised by~\citet{srirag2024evaluatingdialectrobustnesslanguage}, who evaluate dialect-robustness of language models using masked MD-3 conversations.
Finally, our cross-dialect baselines on corpora created using Multi-VALUE and GPT-4 discuss the shortcomings of synthetic datasets for dialect adaptation for dialogues, as also noted in~\citet{faisal2024dialectbench}.

\begin{table}[h!]
    \centering
    \begin{adjustbox}{width=1.0\linewidth,center}
        \begin{tabular}{cc}
        \toprule
         {Feature} & {Example}\\\midrule[\heavyrulewidth]
         \textsc{extraneous article} & \emph{you can combine \underline{the} both the words} \\
         \textsc{lack of inversion in wh-questions} & \emph{what \underline{we can} see in the rivers?}\\
         \textsc{left dislocation} & \emph{\underline{If we have a five sides}, what do we call that?}\\
         \textsc{article omission} & \emph{I'll explain you (the) second word}\\
         \textsc{object fronting} & \emph{\underline{some towers type} it will be}\\
         \textsc{focus} \emph{only} & \emph{I'm trying to explain that \underline{only}}\\
         \textsc{non-initial existential} & \emph{brand names also \underline{there}}\\
         \textsc{mass nouns as count nouns} & \emph{How the women\underline{s} will be?}\\
         \textsc{invariant tag} & \emph{put them on some type of wire \underline{no?}}\\
        \bottomrule
        %& 13.7 & 19.9
        \end{tabular}
    \end{adjustbox}
    \caption{\label{tab:dial_feat} Dialect features identified in erroneously labelled en-IN conversations with the corresponding examples.}
\end{table}
\section{Conclusion}
\label{sec:conclusion}
This paper focused on a simplistic causal language modeling task, called target word prediction, using masked game-playing conversations between two dialectal speakers of English (en-US, en-IN and en-NG). The task was to predict the target word from a masked conversation. From our initial experiments with fine-tuned decoder models, the in-dialect baseline (en-IN and en-NG) reported a performance degradation on TWP, when compared with the skyline (en-US). To address the gap in the case of en-IN and en-NG, we proposed \ours as a novel architecture using low-rank adapters. \ours extends past work in dialect adaptation for encoder models to decoder models by employing contrastive learning via a pseudo-parallel corpus of real conversations. \ours outperformed one in-dialect baseline and three cross-dialect baselines, while also bridging the gap with the skyline to 12\% (down from 27.3\%) and 25\% (down from 64.7\%) on similarity and accuracy respectively for en-IN. For en-NG, the gap is reduced to 5.8\% (down from 17.9\%) on similarity and 4.5\% (down from 43.1\%) on accuracy. Through ablation tests on \ours, we validated the effectiveness of its components. 

Although TWP works with a restricted dataset and utilises turn-based dialogue, \ours sets up the promise for dialect adaptation of decoder models. Our error analysis also highlights the scope for future improvement. A potential future work is to evaluate \ours on other causal language modeling tasks, including seq2seq tasks, and other dialects. Similarly, an extension to \ours would eliminate the requirement of naturally occurring conversations in multiple dialects.% about the same topic.

\section*{Limitations}
While previous approaches have proposed dialect adapters as task-agnostic, our study does not make the same claim. We use target word prediction as the task of predicting the last word of a conversation which was the word that the described was attempting to convey to the guesser. This task is a simplistic version of causal language modeling. However, we do not verify that \ours works for causal language modeling because there is no suitable parallel dataset of turn-aligned conversations, to the best of our knowledge. ~\citet{held-etal-2023-tada} use bottleneck adapters based on their ability for cross-lingual transfer, but we do not explore these types of adapters due to the lack of support for our choice of models at the time of writing the paper. The choice of en-IN and en-NG as the dialects of interest is solely based on the availability of the dataset.

\section*{Ethics Statement}
We use a publicly available dataset of conversations consisting of human players engaged in a game of taboo. The topics discussed in the dataset are fairly general and are unlikely to cause distress. One of the authors of the paper performed the error analysis. The synthetic conversation created using GPT-4 may contain biased output, arising due to the properties of the model. We do not expect any reasonably significant risks arising as a result of the project.

\bibliography{anthology}

\appendix
\onecolumn
\section{Dataset Construction}\label{sec:data_con}
Table~\ref{tab:trans} describes the example conversations from en-IN and en-US subsets along with their respective transformed IN-TR and IN-MV conversations. We utilise the following prompt used in the evaluation study by~\citet{srirag2024evaluatingdialectrobustnesslanguage} to create IN-TR.
\begin{quote}
    \emph{`Normalise the conversation. Remove all exaggerations and dialectal information. Return a neutral response.'}
\end{quote}

The conversations are then masked by replacing the target word with the [MASK] token and pruning the rest of the conversation, as described in the Table~\ref{tab:mask}.

\begin{table*}[h!] 
    \begin{adjustbox}{width=0.8\linewidth,center}
        \renewcommand{\arraystretch}{1}
        \begin{tabular}{p{18em}p{18em}}
            \toprule
             \multicolumn{1}{>{\centering\arraybackslash}m{18em}}{{\IndEng}} & \multicolumn{1}{>{\centering\arraybackslash}m{18em}}{{\Trans}} \\\midrule[\heavyrulewidth]
             \texttt{\textbf{Describer:} \textbf{(}Uh\textbf{)}. What do you call \emph{\underline{\textbf{if we, what will be there}}} in the water?} & \texttt{\textbf{Describer:} \textbf{(}$\varnothing$\textbf{)} What do you call \emph{\underline{\textbf{the creatures}}} in the water?}\\
             \texttt{\textbf{Guesser:} Fish\textbf{(}es\textbf{)}} & \texttt{\textbf{Guesser:} Fish\textbf{(}$\varnothing$\textbf{)}.}\\
             \texttt{\textbf{Describer:} Who \emph{\underline{\textbf{will catch that}}}?} & \texttt{\textbf{Describer:} Who \emph{\underline{\textbf{catches them}}}?}\\
             \texttt{\textbf{Guesser:} Fisherm\emph{\underline{\textbf{a}}}n.} & \texttt{\textbf{Guesser:} Fisherm\emph{\underline{\textbf{e}}}n.}\\\midrule[\heavyrulewidth]
             \multicolumn{1}{>{\centering\arraybackslash}m{18em}}{{\USEng}} & \multicolumn{1}{>{\centering\arraybackslash}m{18em}}{{\Multi}} \\\midrule[\heavyrulewidth]
             \texttt{\textbf{Describer:} Perfect. Oh! \textbf{(}We\textbf{)} earn this. We go to our jobs.} & \texttt{\textbf{Describer:} Perfect. Oh! \textbf{(}$\varnothing$\textbf{)} \textbf{\underline{[are]}} earn\textbf{\underline{[ing]}} this. We \textbf{\underline{[are]}} go\textbf{\underline{[ing]}} to our jobs.}\\
             \texttt{\textbf{Guesser:} Money} & \texttt{\textbf{Guesser:} Money}\\
            \bottomrule
        \end{tabular}
    \end{adjustbox}
    \caption{Example \emph{transformations} of \IndEng to \Trans, and \USEng to \Multi. We utilise GPT-4 Turbo to generate \Trans, and Multi-VALUE to create \Multi. The text in parentheses refers to the omission/removal of certain filler and exaggerated words, and the text such as \texttt{\emph{\textbf{\underline{this}}}}, refers to the words or sentences that were rephrased to convey the original meaning, and the text such as \texttt{\textbf{\underline{[this]}}}, refers to the dialectal features added using Multi-VALUE.}
    \label{tab:trans}
\end{table*}

\begin{table*}[]
    \begin{adjustbox}{width=0.8\linewidth,center}
        \renewcommand{\arraystretch}{1}
        \begin{tabular}{cp{18em}p{18em}}
            \toprule
            \multicolumn{1}{>{\centering\arraybackslash}m{4em}}{{Target Word}} & \multicolumn{1}{>{\centering\arraybackslash}m{18em}}{{\IndEng}} & \multicolumn{1}{>{\centering\arraybackslash}m{18em}}{{Masked \IndEng}} \\\midrule[\heavyrulewidth]
            \multirow{4}{4em}{\centering Fisherman} & \texttt{{Describer:} Uh. What do you call if we, what will be there in the water?} & \texttt{{Describer:} Uh. What do you call if we, what will be there in the water?}\\
            & \texttt{{Guesser:} Fishes} & \texttt{{Guesser:} Fishes}\\
            & \texttt{{Describer:} Who will catch that?} & \texttt{{Describer:} Who will catch that?}\\
            & \texttt{{Guesser:} \emph{\underline{\textbf{Fisherman}}}.} & \texttt{{Guesser:} \emph{\underline{\textbf{[MASK]}}}}\\\midrule[\heavyrulewidth]
            \multicolumn{1}{>{\centering\arraybackslash}m{4em}}{{Target Word}} & \multicolumn{1}{>{\centering\arraybackslash}m{18em}}{{\USEng}} & \multicolumn{1}{>{\centering\arraybackslash}m{18em}}{{Masked \USEng}} \\\midrule[\heavyrulewidth]
            \multirow{3}{4em}{\centering Planet} & \texttt{{Describer:} These are hard words. um Okay. So there's. the Sun and the Moon and all the rest of them.} & \texttt{{Describer:} These are hard words. um Okay. So there's. the Sun and the Moon and all the rest of them.}\\
            & \texttt{Guesser: And all the \emph{\underline{\textbf{planet}}}s?} & \texttt{Guesser: \emph{\underline{\textbf{[MASK]}}}}\\
            & \texttt{\textbf{(}Describer: Yes.\textbf{)}} & \\
            \bottomrule
        \end{tabular}
    \end{adjustbox}
    \caption{Masking conversations from the extended MD-3. The text such as \texttt{\emph{\textbf{\underline{this}}}} represents the target word utterance by the guesser which is masked (represented by, \texttt{\emph{\textbf{\underline{[MASK]}}}} in the final version of the conversation. The rest of the original conversation is pruned as represented text in parentheses.}
    \label{tab:mask}
\end{table*}

\begin{table*}[] 
    \begin{adjustbox}{width=0.8\linewidth,center}
        \renewcommand{\arraystretch}{1}
        \begin{tabular}{cp{18em}p{18em}}
            \toprule
              Label & \multicolumn{1}{>{\centering\arraybackslash}m{18em}}{{\USEng}} & \multicolumn{1}{>{\centering\arraybackslash}m{18em}}{{\IndEng}} \\\midrule[\heavyrulewidth]
             \multirow{2}{4em}{Positive}& \texttt{\textbf{Describer:} Good job. Okay. Um. How we. How we clean our clothes.} & \texttt{\textbf{Describer:} Yeah here I got a thing uh which most of us daily use that to wash our clothes.}\\
             & \texttt{\textbf{Guesser:} [MASK]} & \texttt{\textbf{Guesser:} [MASK]}\\\hdashline
             \multirow{4}{4em}{Negative} & \texttt{\textbf{Describer:} this. What? All right all right so.} & \texttt{\textbf{Describer:} Yeah here I got a thing uh which most of us daily use that to wash our clothes.}\\
             & \texttt{\textbf{Guesser:} What?} & \texttt{\textbf{Guesser:} [MASK]}\\
             & \texttt{\textbf{Describer:} Uh this uh this young man. um is a very well-known singer. who was kind of a heart-throb. Hm he I mean he's still active but like 10 years ago like all of the girls were crazy about this guy.} & \\
             & \texttt{\textbf{Guesser:} [MASK]} & \\
            \bottomrule
        \end{tabular}
    \end{adjustbox}
    \caption{Example conversation pairs from the pseudo-parallel corpus: en-US || en-IN. A positive example contains conversations describing the same target word, while the negative example contains conversations pertaining to two different target words.}
    \label{tab:neg_pos}
\end{table*}

Table~\ref{tab:neg_pos} describes examples from the pseudo-parallel corpus: en-US || en-IN. The conversations in a positive pair, while dissimilar in the syntax of the conversation, pertain to the same target word. For example, the conversation pair labelled as `positive' in the Table~\ref{tab:neg_pos} describe the same target word-- \emph{Washing Machine}. The conversation pair labelled as `negative' describe different target words; the en-US conversation describes \emph{Justin Bieber}, while en-IN conversation describes \emph{Washing Machine}.

\section{Additional Ablations}\label{sec:add_ab}
We conducted additional ablation studies on \ours to address the following question: Can the performance improvement of \ours be attributed to the increased training data from data augmentation?

Table~\ref{tab:add_ab_1} compares the performance of the proposed combination of \ours with variations that exclude data augmentation. Training the task adapter solely on en-IN results in significantly lower performance, with similarity scores dropping by 5.9 to 7.0 and accuracy scores decreasing by 8.2 to 9.7.

Table~\ref{tab:add_ab_2} examines the effect of varying the proportion of en-US conversations in the augmented training data (en-US + en-IN). The best performance is observed when \ours is trained with augmented data containing only 50\% en-US conversations. While this configuration outperforms the proposed full-proportion combination, determining the optimal proportions is challenging and limits generalisability across models. More particularly, Table~\ref{tab:add_ab_2} also reveals that \mistral is highly sensitive to such changes in the training data composition, whereas \gemma is more robust.

These ablation results, combined with the findings in Table~\ref{tab:ablation}, further reinforce our proposed methodology. Specifically, training the task adapter on fully proportioned augmented data (en-IN + en-US) and the dialect adapter on a parallel corpus constructed from natural conversations (en-US || en-IN) proves to be a more effective and generalisable approach.
\begin{table*}[]
    \centering
    \begin{adjustbox}{width=1.0\linewidth,center}
        \begin{tabular}{m{5em}m{6.68em}m{6.68em}cccccc}
        \toprule
        \multirow{2}{5em}{\centering Method} & \multirow{2}{6.68em}{\centering Training Data} & \multirow{2}{6.68em}{\centering $\textbf{||}_{\text{Corpus}}$}& \multicolumn{2}{c}{\mistral} & \multicolumn{2}{c}{\gemma} & \multicolumn{2}{c}{$\mu$}\\\cmidrule(lr){4-5}\cmidrule(lr){6-7}\cmidrule(lr){8-9}
          & & & Similarity & Accuracy & Similarity & Accuracy & Similarity & Accuracy\\\midrule[\heavyrulewidth]
         \centering \ours %
         & \centering en-US + en-IN & \centering en-US || en-IN & \textbf{55.9} & \textbf{30.0} & \textbf{63.9} & \textbf{41.3} & \textbf{59.9} & \textbf{35.7}\\\hdashline
         \multirow{3}{5em}{\centering $\leftrightarrow \textbf{||}_{\text{Corpus}}$} %
         & \multirow{3}{6.68em}{\centering en-IN \small{(No Augmentation)}} & \centering en-US || en-IN & 52.0 & 23.1 & 53.7 & 28.8 & 52.9 \diff{7.0} & 26.0 \diff{9.7}\\
         & & \centering en-IN || IN-TR & 52.0 & 23.8 & 54.1 & 28.8 & 53.0 \diff{6.9} & 26.3 \diff{9.4}\\
         & & \centering en-US || IN-MV & 53.3 & 25.0 & 54.6 & 30.0 & 54.0 \diff{5.9} & 27.5 \diff{8.2}\\
        \bottomrule
        \end{tabular}
    \end{adjustbox}
    \caption{\label{tab:add_ab_1}
    Ablation on \ours based on parallel corpus ($\leftrightarrow \textbf{||}_{\text{Corpus}}$) and data augmentation. For each model, we report Similarity and Accuracy when tested on en-IN. The best performance is shown in \textbf{bold}. $\mu$ is the average of the metrics across both models. The degradation on the ablations compared to \ours is shown in \diff{number}.
    }
\end{table*}

\begin{table*}[]
    \centering
    \begin{adjustbox}{width=1.0\linewidth,center}
        \begin{tabular}{m{5em}m{6.68em}m{6.68em}cccccc}
        \toprule
        \multirow{2}{5em}{\centering Method} & \multirow{2}{6.68em}{\centering $\textbf{||}_{\text{Corpus}}$} & \multirow{2}{6.68em}{\centering \% of en-US} & \multicolumn{2}{c}{\mistral} & \multicolumn{2}{c}{\gemma} & \multicolumn{2}{c}{$\mu$}\\\cmidrule(lr){4-5}\cmidrule(lr){6-7}\cmidrule(lr){8-9}
          & & & Similarity & Accuracy & Similarity & Accuracy & Similarity & Accuracy\\\midrule[\heavyrulewidth]
         \multirow{5}{5em}{\centering \ours} & \multirow{5}{6.68em}{\centering en-US || en-IN} %
            & \centering 0\% & 52.0 & 23.1 & 53.7 & 28.8 & 52.9 & 26.0\\
            & & \centering 25\% & 53.8 & 31.9 & 61.2 & 35.4 & 57.5 & 33.7\\
            & & \centering 50\% & \textbf{58.8} & \textbf{33.8} & \textbf{64.1} & \textbf{41.8} & \textbf{61.5} & \textbf{37.8}\\
            & & \centering 75\% & 54.6 & 30.6 & 63.4 & 40.8 & 59.0 & 35.7\\
            & & \centering 100\%* & 55.9* & 30.0* & 63.9* & 41.3* & 59.9* & 35.7*\\\hdashline
         \multirow{5}{5em}{\centering $-\mathcal{L}_{\texttt{\textbf{Dial}}}$} & \multirow{5}{6.68em}{\centering Not Used} %
            & \centering 0\% & 51.0 & 24.4 & 54.6 & 30.0 & 52.8 & 27.2\\
            & & \centering 25\% & 52.0 & 29.4 & 60.5 & 34.4 & 56.3 & 31.9\\
            & & \centering 50\% & 55.3 & 29.4 & 61.4 & 35.6 & 58.4 & 32.2\\
            & & \centering 75\% & 52.5 & 27.5 & 61.6 & 35.6 & 57.1 & 31.6\\
            & & \centering 100\% & 54.4 & 26.9 & 62.3 & 37.5 & 58.4 & 32.2\\
        \bottomrule
        \end{tabular}
    \end{adjustbox}
    \caption{\label{tab:add_ab_2}
    Ablation on \ours based on dialect adapter ($\mathcal{L}_{\texttt{\textbf{Dial}}}$) and proportion of en-US conversations in augmented data (en-US + en-IN). For each model, we report Similarity and Accuracy when tested on en-IN. The best performance is shown in \textbf{bold}, and the proposed combination is represented by number*. $\mu$ is the average of the metrics across both models.
    }
\end{table*}

\end{document}